\documentclass{article}

\usepackage{arxiv}

\usepackage[utf8]{inputenc} 
\usepackage[T1]{fontenc}    
\usepackage{hyperref}       
\usepackage{url}            
\usepackage{booktabs}       
\usepackage{amsfonts}       
\usepackage{nicefrac}       
\usepackage{microtype}      
\usepackage{lipsum}		
\usepackage{graphicx}
\usepackage[numbers]{natbib}
\usepackage{doi}
\usepackage{caption}
\usepackage{threeparttable}
\usepackage{placeins}
\usepackage{float} 
\usepackage[affil-it]{authblk} 
\usepackage{hyperref} 

\title{Is open-source there yet? A comparative study on commercial and open-source LLMs in their ability to label Chest X-Ray reports}

\author[1,2]{\textbf{Felix J. Dorfner} \textsuperscript{*}}
\author[3,4]{Liv Jürgensen}
\author[2]{Leonhard Donle}
\author[2]{Fares Al Mohamad}
\author[1]{Tobias R. Bodenmann}
\author[1]{Mason C. Cleveland}
\author[2]{Felix Busch}
\author[5]{Lisa C. Adams}
\author[6]{James Sato}
\author[6]{Thomas Schultz}
\author[1]{Albert E. Kim}
\author[7]{Jameson Merkow}
\author[8]{\textbf{Keno K. Bressem}\textsuperscript{$\ddagger$}}
\author[1,6]{\textbf{Christopher P. Bridge}\textsuperscript{$\dagger$}\textsuperscript{$\ddagger$}}

\affil[1]{Athinoula A. Martinos Center for Biomedical Imaging, Massachusetts General Hospital and Harvard Medical School, 149 Thirteenth St, Charlestown, MA 02129, USA}
\affil[2]{Department of Radiology, Charité - Universitätsmedizin Berlin corporate member of Freie Universität Berlin and Humboldt Universität zu Berlin, Hindenburgdamm 30, 12203 Berlin, Germany}
\affil[3]{Department of Pediatric Oncology, Dana-Farber Cancer Institute, Boston, MA, USA}
\affil[4]{Charité - Universitätsmedizin Berlin, Corporate member of Freie Universität Berlin and Humboldt-Universität zu Berlin, 10117 Berlin, Germany}
\affil[5]{Department of Diagnostic and Interventional Radiology, Technical University of Munich, Munich, Germany}
\affil[6]{Mass General Brigham Data Science Office, Boston, MA, USA}
\affil[7]{Microsoft Health and Life Sciences (HLS), Redmond, WA, USA}
\affil[8]{Department of Radiology and Nuclear Medicine, German Heart Center Munich, Munich, Germany}

\date{}

\renewcommand{\headeright}{}
\renewcommand{\shorttitle}{Large Language Models for x-ray report labeling}

\hypersetup{
pdftitle={Is open-source there yet? A comparative study on commercial and open-source LLMs in their ability to label Chest X-Ray reports},
pdfsubject={Chest X-Ray, Large Language Models, Artificial Intelligence, Natural Language Processing, Radiology},
pdfauthor={Felix J. Dorfner, Liv Jürgensen, Leonhard Donle, Fares Al Mohamad, James Sato, Tom Schulz, Tobias R. Bodenmann, Mason C. Cleveland, Felix Busch, Lisa C. Adams  Albert E. Kim, Jameson Merkow, Keno K. Bressem, Christopher P. Bridge},
pdfkeywords={Chest X-Ray, Large Language Model, Artificial Intelligence},
}

\begin{document}
\maketitle
\textsuperscript{*}First author.\\
\textsuperscript{$\dagger$}Corresponding author: cbridge@mgh.harvard.edu \\
\textsuperscript{$\ddagger$}These authors contributed equally as last authors.

\abstract{\textbf{Introduction:} With the rapid advances in large language models (LLMs), there have been numerous new open source as well as commercial models. While recent publications have explored GPT-4 in its application to extracting information of interest from radiology reports, there has not been a real-world comparison of GPT-4 to different leading open-source models.
 
\textbf{Materials and Methods:} Two different and independent datasets were used. The first dataset consists of 540 chest x-ray reports that were created at the Massachusetts General Hospital between July 2019 and July 2021. The second dataset consists of 500 chest x-ray reports from the ImaGenome dataset. We then compared the commercial models GPT-3.5 Turbo and GPT-4 from OpenAI to the open-source models Mistral-7B, Mixtral-8x7B, Llama2-13B, Llama2-70B, QWEN1.5-72B and CheXbert and CheXpert-labeler in their ability to accurately label the presence of multiple findings in x-ray text reports using different prompting techniques. 

\textbf{Results:} On the ImaGenome dataset, the best performing open-source model was Llama2-70B with micro F1-scores of 0.972 and 0.970 for zero- and few-shot prompts, respectively. GPT-4 achieved micro F1-scores of 0.975 and 0.984, respectively. On the institutional dataset, the best performing open-source model was QWEN1.5-72B with micro F1-scores of 0.952 and 0.965 for zero- and few-shot prompting, respectively. GPT-4 achieved micro F1-scores of 0.975 and 0.973, respectively.

\textbf{Conclusion:} In this paper, we show that while GPT-4 is superior to open-source models in zero-shot report labeling, the implementation of few-shot prompting can bring open-source models on par with GPT-4. This shows that open-source models could be a performant and privacy preserving alternative to GPT-4 for the task of radiology report classification.}

\endabstract

\section{Introduction}\label{sec1}

Over the past year, Large Language Models (LLMs) have become increasingly popular and have had a tremendous impact on the field of natural language processing \cite{thirunavukarasu_large_2023, zhao_survey_2023}. They have demonstrated the ability to perform a wide range of tasks in various domains that were previously exclusive to humans. In medicine and radiology in particular, there have been a number of applications of LLMs, such as using them to transform radiology reports into a structured reporting format \cite{adams_leveraging_2023}, using a LLM for various classification and summarization tasks for radiology reports \cite{liu_exploring_2023}, and using various closed-source LLMs to classify radiology reports for the presence of urgent or emergency findings \cite{infante_large_2024}. By extracting information at scale from the unstructured, free-text data available, these models could enable studies to be conducted on a scale previously considered impractical. This has applications in a variety of research areas, including epidemiological studies and training of artificial intelligence (AI) models.
 
Despite these advances, as in the case of the papers cited above, much of the attention has been focused on commercial LLMs such as OpenAI's GPT-4. While this model is undoubtedly capable, there are several drawbacks to using a commercial, closed-source model. First, using GPT-4 and other proprietary models requires sending data to remote servers where the model is running. This raises privacy concerns, which are particularly critical for medical text reports that contain highly sensitive patient information. The associated regulatory restrictions vary in different parts of the world, but may make this impossible or challenging for many healthcare researchers. Second, the use of proprietary models requires communication with the external server through an application programming interface (API). These interfaces are regularly updated and changed, which can render code previously used to send input to the model obsolete. Third, not only the interfaces but also the models are updated, and old versions may no longer be available, creating problems in research applications where consistency and reproducibility are essential. Fourth, the use of OpenAI's models through the API is charged on a per token basis, which can incur significant costs when used extensively.

Freely available LLMs such as Llama, Mistral or QWEN do not have these limitations, since the models can be used locally, ensuring both privacy and reproducibility along with a substantial reduction in direct costs to the models. Despite these advantages, it is still unclear how performance of these open source models compares to live service models such as GPT-4 in real-world medical tasks. In this work, we aim to explore the capabilities of popular open-source models on a specific radiology task and compare their performance with GPT-4, the current state of the art for LLMs.

We focus on classifying full-text radiology reports for a list of important findings, an example of unstructured text to structured data transformation. Currently, the generation of labels from radiology reports is an expert-driven and labor-intensive task. Consequently, a significant portion of medical AI studies depend on a limited selection of publicly accessible large datasets or considerably smaller datasets from individual institutions.
Automatically deriving labels from unstructured radiology reports presents an efficient solution to address these challenges. To automate this process, we must use models that understand medical text, including highly specialized language and nuanced expressions of uncertainty, an obstacle exacerbated by variation in formatting and language used at different institutions. LLMs stand out as a potential solution to this challenge since they excel at transforming free-text into structured labels and are able to utilize few-shot learning techniques to avoid costly retraining procedures. 

Current research is predominantly focused on GPT-4, despite the existence of numerous open-source models. These models, despite their potential, remain largely overlooked, with their performance metrics and capabilities yet to be comprehensively explored or understood.
In this work, we compare GPT-3.5 and GPT-4 with the open-source models Llama2-13B, Llama2-70B, Mistral-7B, Mixtral-8x7B, QWEN1.5-72B in their ability to classify radiology reports. As a baseline, we use CheXbert, the current state of the art in radiology report classification, and CheXpert-labeler.

\section{Methods}\label{sec2}

\subsection{Ethics approval}
The institutional review board approved this Health Insurance Portability and Accountability Act–compliant, retrospective secondary analysis under IRB 2022P002646. The need for informed consent was waived. Access to GPT-3.5 and GPT-4 was through a HIPAA-compliant cloud implementation.

\subsection{Cohort Description}
Radiology reports from chest x-ray examinations from two different datasets were used.
The first dataset was created using the gold level annotations that the ImaGenome dataset \cite{wu_chest_2021, goldberger_physiobank_2000} provides for 500 radiology reports in the MIMIC-CXR Database \cite{johnson_mimic-cxr_2019, johnson_mimic-cxr_2019-1}. Of the 500 reports, 50 were randomly chosen as a subset for drawing the few-shot reports. These reports were subsequently excluded from the analysis. 

The second dataset consists of radiology reports for chest x-rays that were created at the Massachusetts General Hospital between July 2019 and July 2021. 540 reports were selected from this dataset. Of the 540 reports, 40 were randomly chosen as a subset for drawing the few-shot reports. These reports were subsequently excluded from the analysis.

\begin{figure}
    \centering
    \includegraphics[width=0.5\linewidth]{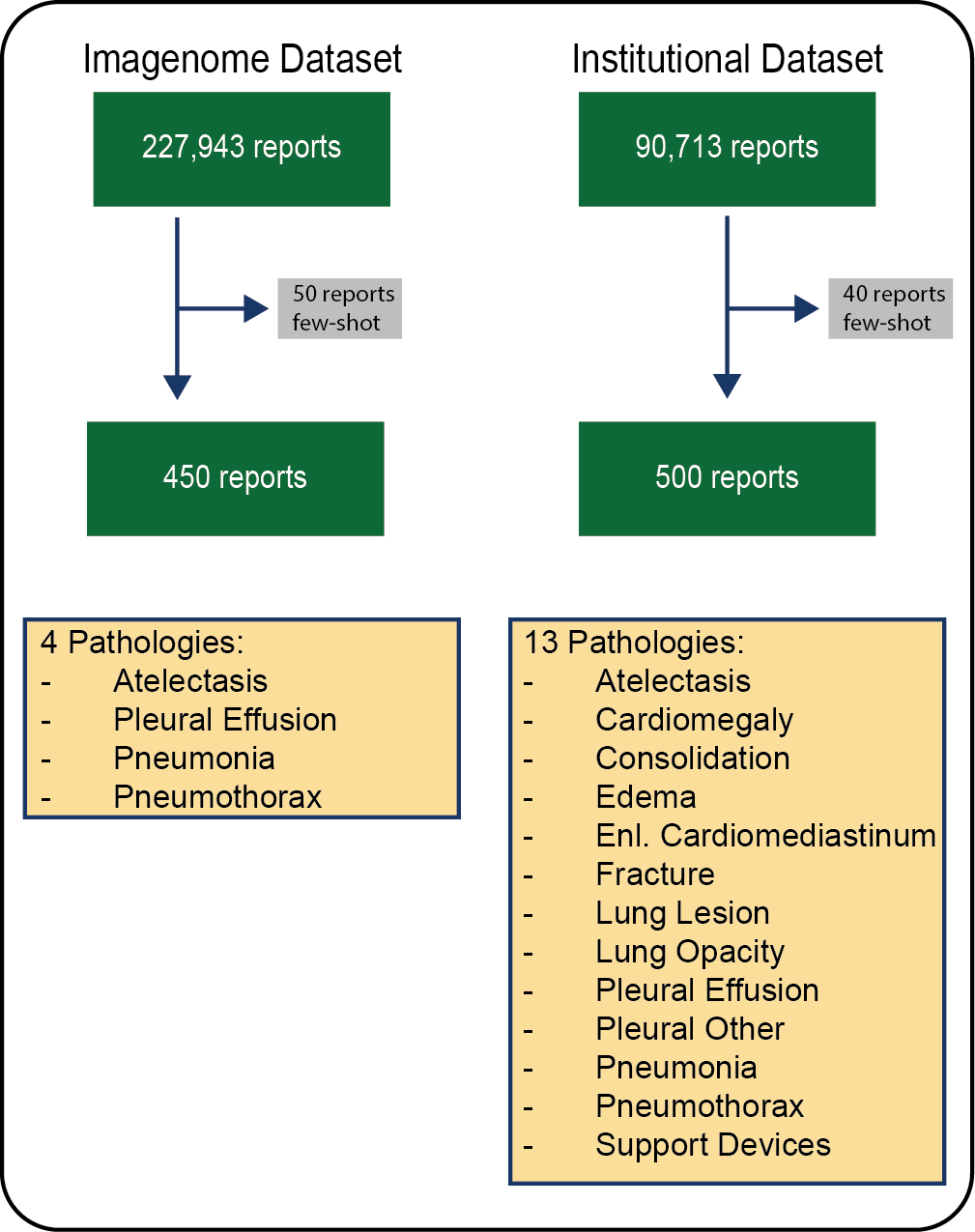}
    \caption{Flow-chart for the reports from the ImaGenome and the institutional dataset}
    \label{fig:patient_flow}
\end{figure}

\subsection{Label Generation}
The ImaGenome Gold dataset was made publicly available by Wu et al. on PhysioNet. It provides labels that were first automatically generated and then manually evaluated and corrected by experts for 500 chest x-ray reports. This was done by physicians with 2 to 10 or more years of clinical experience. One of the annotators was a U.S.-trained radiologist with more than 6 years of radiology experience. The ImaGenome dataset was annotated on a sentence basis. For our work, we combined the sentence-based labels at the report level and paired them with their corresponding radiology reports from the MIMIC-CXR database, creating binary labels for the presence/absence of atelectasis, pleural effusion, pneumonia, and pneumothorax. 

The institutional dataset was labeled independently by 2 fourth-year medical students. In case of disagreement of the 2 annotators, the case was settled by a board-certified radiologist, with 7 years of experience in radiology. The reports were labeled according to the CheXpert-labeler rules, and one of three possible labels: ``Yes'', ``No'' and ``Maybe'' were assigned for each of the 13 findings as listed in Figure \ref{fig:patient_flow}.

\subsection{LLM-based Report Labeling}
The models listed in Table \ref{tab:model_comparison} were used for the task of report classification. In addition to the generalist models, the current state of the art model for chest x-ray report labeling CheXbert and the rule-based CheXpert-labeler were used \cite{smit_CheXbert_2020, irvin_chexpert_2019}. The models were implemented using vLLM \cite{kwon_efficient_2023} and prompted using the chat completion interface. We observed that Mixtral-8x7B's answers were often cut short under this approach, resulting in invalid JSON, so we instead used the generation interface for that model. To reduce the memory footprint, the Llama-2-70b and QWEN-1.5-72B models were run in their Int4 quantized version using Activation-aware Weight Quantization (AWQ) \cite{lin_awq_2023}.  We decided to use the ``Instruct'' or ``Chat'' versions of the models, which have been additionally fine-tuned on Instruction and Chat Completions tasks, as they are better suited to our use case. The model parameters such as temperature, top\_p and penalties were the same for all models. The temperature was set to zero, to ensure outputs that are as close to deterministic as possible. The top\_p value was set to 0.95, which is the default setting for the OpenAI API. The frequency and repetition penalty were set to zero for our task, where label tokens are frequently repeated in the model output. Additionally, we provide results for an ensemble model, which was constructed by combining the predictions of Mixtral-8x7B, Llama2-70B and QWEN1.5-72B through a majority vote.

All models were provided with identical prompts, employing two popular prompting techniques: zero-shot and few-shot prompting \cite{brown_language_2020}. In zero-shot experiments, the prompts consisted of a brief task description along with a template outlining the desired JSON format for the response. Few-shot experiments used the same task description and template, supplemented by examples included in the prompt. These examples consisted of reports paired with their corresponding desired output labels. Experiments were conducted on a development set to find the ideal strategies for choosing the examples. We then chose the best performing strategy for selecting the few-shot examples, which was to randomly draw reports to include at least one positive label for each finding while using as few examples as possible. This approach yielded two example reports for the ImaGenome dataset and four example reports for the institutional dataset.
Additional post-processing was applied to the results to conform to the CheXbert standard and provide a basis for fair evaluation. The CheXpert-labeler rules have a hierarchical structure, wherein any report labeled positive for Cardiomegaly would also be labeled positive for Enlarged Cardiomediastinum. Similarly, the reports labeled positive for any of the findings Edema, Consolidation, Pneumonia, Lung Lesion and Atelectasis are automatically labeled positive for Lung Opacity. The CheXpert-labeler tasks use four labels: ``Yes'', ``Maybe'', ``No'' and ``No Information''. With the downstream task of creating labels for training vision models, we merged the ``No Information''/``Undefined'' and ``No'' labels into ``No''. For the Imagenome datasets, the available labels are ``Yes'' and ``No''. For the institutional dataset, manual annotations were created for the labels ``Yes'', ``Maybe'', and ``No''. Experiments were run on all three labels for the institutional dataset and as a binary classification task where the ``Maybe'' labels provided by the models were converted to ``Yes'' for both datasets.

\begin{figure}
    \centering
    \includegraphics[width=1\linewidth]{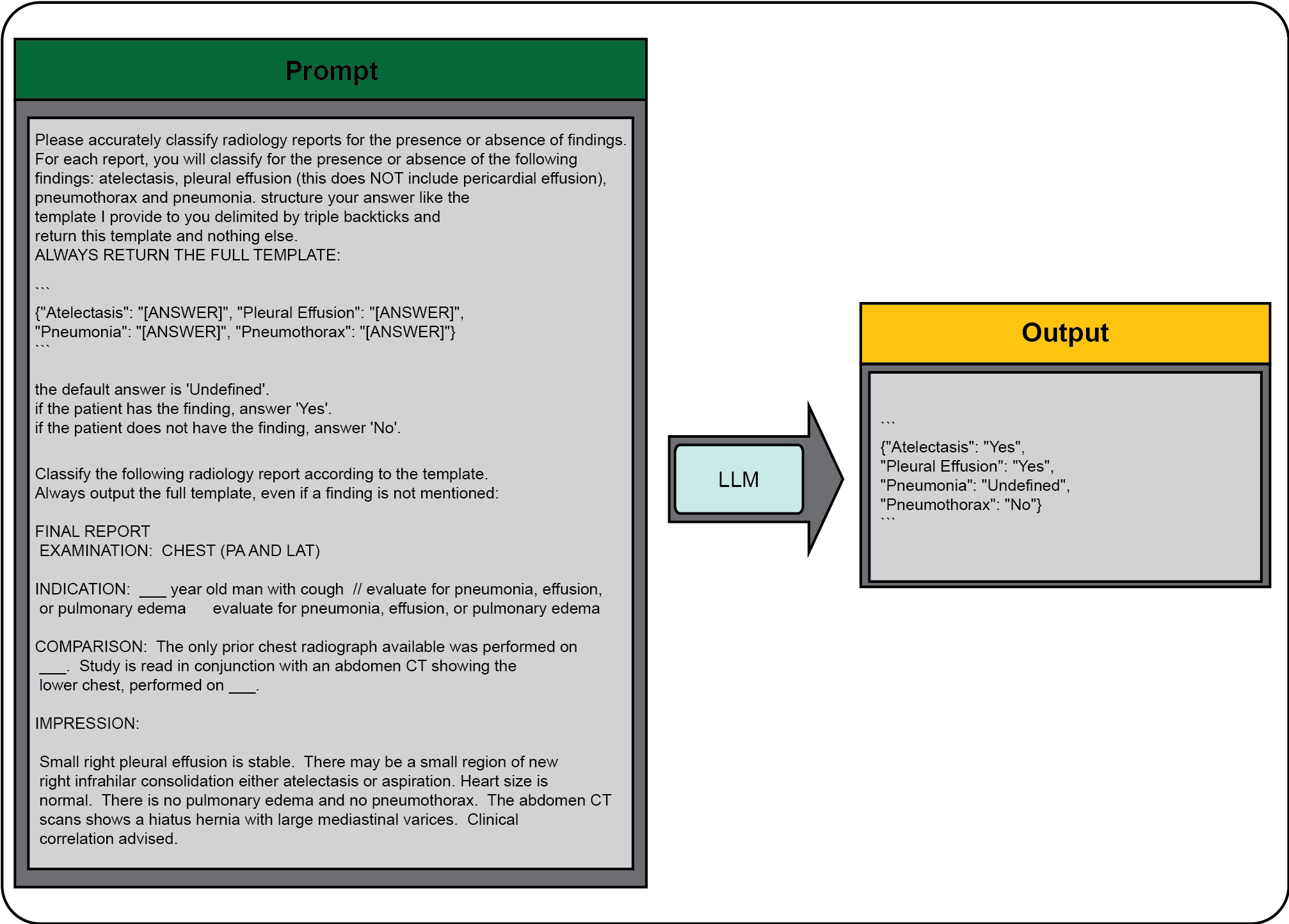}
    \caption{Outline of the zero-shot prompting workflow for the ImaGenome dataset. The model is given instructions and the report to classify and outputs its response according to the JSON formatted template with the corresponding labels.}
    \label{fig:prompt}
\end{figure}

\begin{table}[ht]
\centering
\caption{Comparison of language models with respect to their parameters, context length, special features, release date, and publisher.}
\begin{threeparttable}
\begin{tabular}{lcccc}
\textbf{Model} & \textbf{Parameters} & \textbf{Context Length} & \textbf{Release Date} & \textbf{Publisher} \\
\hline
Llama-2-13b-chat \cite{touvron_llama_2023} & 13B & 4k & July 18, 2023 & Meta \\
Llama-2-70b-chat \cite{touvron_llama_2023} & 70B & 4k & July 18, 2023 & Meta \\
Mistral-7B-Instruct-v0.2 \cite{jiang_mistral_2023} & 7B & 8k & December 11, 2023 & Mistral AI \\
Mixtral-8x7B-Instruct-v0.1 \cite{jiang_mixtral_2024} & 46.7B/12.9B\tnote{1} & 32k & December 11, 2023 & Mistral AI \\
Qwen1.5-72B-Chat \cite{bai_qwen_2023} & 72B & 32k & February 4, 2024 & Alibaba Cloud \\
GPT-3.5 Turbo & not disclosed\tnote{2} & 16k & March 1, 2023 & OpenAI \\
GPT-4 \cite{openai_gpt-4_2023} & not disclosed\tnote{2} & 32k & March 14, 2023 & OpenAI \\

\hline
\end{tabular}
\begin{tablenotes}
\item[1] As a sparse mixture-of-experts model (8 ``experts''), it has a total of 46.7B total parameters but only uses 12.9B parameters (2 experts) per token.
\item[2] GPT-3.5 Turbo and GPT-4 are proprietary models; the model weights are not publicly available.
\end{tablenotes}
\end{threeparttable}
\label{tab:model_comparison}
\end{table}

\subsection{Statistical Analysis}
The Python packages NumPy (Version 1.26.3), pandas (Version 2.1.4), scikit-learn (Version 1.4.0), statsmodels (Version 0.14.1), matplotlib (Version 3.8.2), and seaborn (Version 0.13.2) were used for data analysis and visualization \cite{harris_array_2020, mckinney_data_2010, pedregosa_scikit-learn_2011, hunter_matplotlib_2023, seabold_statsmodels_2010, waskom_seaborn_2021}. McNemar's test was used to compare the models' performance to GPT-4 \cite{dietterich_approximate_1998}. Bonferroni's correction for multiple comparisons was used. A $p$-value of less than 0.05 was considered statistically significant. Given data imbalances across datasets, micro and macro F1-Scores were calculated.

\section{Results}\label{sec3}
\subsection{Imagenome Dataset}

The ImaGenome dataset consists of 450 radiology reports after removing the subset used to draw few-shot examples. The number of pathological findings for each class in the dataset can be found in the appendix (Table \ref{tab:findings_imagenome}). 

For the first experiment, we compared the models using a zero-shot prompt. This prompt provided instructions on how to complete the task and a template for the expected JSON output. Table \ref{tab:zero_shot_results_imagenome} shows the results of this experiment. The results achieved by CheXpert-labeler and CheXbert are also shown in the table for comparison. For CheXpert-labeler and CheXbert the label ``Maybe'' was collapsed into ``Yes'', to allow comparison on this dataset. It is important to note here that those models do not take a prompt input, they were used according to their instructions and provided with the same text reports in their desired input format. GPT-4 was the best performing model on this task with a Micro F1-Score of 0.975. Llama2-70B had a very similar performance of 0.972. Performance varied across the different labels, with Pneumonia being the worst performing finding overall.

\begin{table}[ht]
\caption{F1-scores for zero-shot prompting on the ImaGenome dataset}
\centering
\begin{threeparttable}
\resizebox{\textwidth}{!}{
\begin{tabular}{lcccccccccc}
\textbf{Finding} & \textbf{CheXpert}\tnote{**}\hspace{8pt}\tnote{1}& \textbf{CheXbert}\tnote{*}\hspace{4pt}\tnote{1}& \textbf{Mistral-7B}\tnote{**} & \textbf{Llama2-13B}\tnote{**} & \textbf{Mixtral-8x7B}\tnote{**}& \textbf{Llama2-70B} &\textbf{QWEN1.5-72B}\tnote{**}& \textbf{Ensemble}\tnote{*}\hspace{4pt}\tnote{2}& \textbf{GPT-3.5}\tnote{**} & \textbf{GPT-4} \\
\hline
Atelectasis & 0.963 & 0.964 & 0.942 & 0.876 & 0.942 & 0.956 & 0.928 & 0.96 & 0.824 & 0.982 \\ 
Pleural Effusion & 0.923 & 0.912 & 0.927 & 0.862 & 0.927 & 0.948 & 0.901 & 0.942 & 0.875 & 0.935 \\ 
Pneumonia & 0.632 & 0.759 & 0.418 & 0.606 & 0.418 & 0.783 & 0.500 & 0.602 & 0.621 & 0.777 \\ 
Pneumothorax & 0.462 & 0.783 & 0.875 & 0.800 & 0.875 & 0.824 & 0.875 & 0.875 & 0.800 & 0.941 \\ 
\specialrule{1pt}{0pt}{0pt}
Micro F1 & 0.931 & 0.959 & 0.953 & 0.946 & 0.953 & 0.972 & 0.943 & 0.964 & 0.941 & 0.975 \\ 
Macro F1 & 0.745 & 0.854 & 0.790 & 0.786 & 0.790 & 0.878 & 0.801 & 0.845 & 0.780 & 0.909 \\ 
\specialrule{1pt}{0pt}{0pt}
\end{tabular}
}
\begin{tablenotes}
\scriptsize
\item[1] CheXpert-labeler and CheXbert do not take prompts, but are still listed here for reference 
\item[2] The Ensemble is constructed by combining the predictions of Mixtral-8x7B, Llama2-70B and QWEN1.5-72B through a majority vote. 
\item[*] There was a statistically significant difference in the performance compared to GPT-4 (p<0.05)
\item[**] There was a statistically significant difference in the performance compared to GPT-4 (p<0.01)
\end{tablenotes}

\end{threeparttable}
\label{tab:zero_shot_results_imagenome}
\end{table}

Table \ref{tab:few_shot_results_imagenome} shows the results for few-shot prompting on the ImaGenome dataset. A total of two example reports were provided to the models, which included at least one positive sample for each of the four findings.  GPT-4 was the best performing model on this task with a micro F1-Score of 0.984, which constitutes a 0.9 percentage point improvement from its zero-shot result. Llama2-70B was still the second best performing model with a micro F1-Score of 0.970, but its performance did not improve through the use of few-shot prompting. 

\begin{table}[ht]
\caption{F1-scores for few-shot prompting using one positive example for each finding on the ImaGenome dataset}
\centering
\begin{threeparttable}
\resizebox{\textwidth}{!}{
\begin{tabular}{lcccccccccc}

\textbf{Finding} & \textbf{CheXpert}\tnote{**}\hspace{8pt}\tnote{1}& \textbf{CheXbert}\tnote{**}\hspace{8pt}\tnote{1}& \textbf{Mistral-7B}\tnote{**} & \textbf{Llama2-13B}\tnote{**} & \textbf{Mixtral-8x7B}\tnote{**}& \textbf{Llama2-70B}\tnote{**} &\textbf{QWEN1.5-72B}\tnote{**}& \textbf{Ensemble}\tnote{**}\hspace{8pt}\tnote{2}& \textbf{GPT-3.5}\tnote{**} & \textbf{GPT-4} \\
\hline
Atelectasis & 0.963 & 0.964 & 0.936 & 0.923 & 0.936 & 0.943 & 0.940 & 0.953 & 0.959 & 0.982 \\ 
Pleural Effusion & 0.923 & 0.912 & 0.933 & 0.928 & 0.933 & 0.947 & 0.921 & 0.953 & 0.963 & 0.958 \\ 
Pneumonia & 0.632 & 0.759 & 0.468 & 0.806 & 0.468 & 0.765 & 0.698 & 0.679 & 0.777 & 0.882 \\ 
Pneumothorax & 0.462 & 0.783 & 0.875 & 0.737 & 0.875 & 0.941 & 0.875 & 0.875 & 0.875 & 1.0 \\ 
\specialrule{1pt}{0pt}{0pt}
Micro F1 & 0.931 & 0.959 & 0.954 & 0.961 & 0.954 & 0.97 & 0.961 & 0.967 & 0.974 & 0.984 \\ 
Macro F1 & 0.745 & 0.854 & 0.803 & 0.848 & 0.803 & 0.899 & 0.859 & 0.865 & 0.894 & 0.955 \\ 
\specialrule{1pt}{0pt}{0pt}
\end{tabular}
}
\begin{tablenotes}
\scriptsize
\item[1] CheXpert-labeler and CheXbert do not take prompts, but are still listed here for reference 
\item[2] The Ensemble is constructed by combining the predictions of Mixtral-8x7B, Llama2-70B and QWEN1.5-72B through a majority vote. 
\item[*] There was a statistically significant difference in the performance compared to GPT-4 (p<0.05)
\item[**] There was a statistically significant difference in the performance compared to GPT-4 (p<0.01)
\end{tablenotes}
\end{threeparttable}

\label{tab:few_shot_results_imagenome}
\end{table}

\subsection{Institutional Dataset}

The institutional dataset consists of 500 radiology reports after removing the subset used to draw few-shot examples. The number of pathological findings for each class in the dataset can be found in the appendix (Table \ref{tab:findings_institutional}).

The results of the first experiment which compared the models on the zero-shot prompt can be found in Table \ref{tab:zero_shot_binary_institutional}. For this comparison, the model Outputs ``Yes'' and ``Maybe'' were collapsed to ``Yes'' to create a binary classification case. GPT-4 was the best performing model on this task with a micro F1-Score of 0.975. The second best performing model is QWEN1.5-72B with a micro F-1 Score of 0.952. Llama2-70B achieved a very similar performance of 0.950 on this task. The ensemble consisting of Mixtral-8x7B, Llama2-70B and QWEN1.5-72B achieved a micro F1-score of 0.960.

\begin{table}[ht]
\caption{F1-scores for zero-shot prompting on the Institutional Dataset for the binary case.}
\centering
\begin{threeparttable}
\resizebox{\textwidth}{!}{
\begin{tabular}{lcccccccccc}
\textbf{Finding} & \textbf{CheXpert}\tnote{**}\hspace{8pt}\tnote{1}& \textbf{CheXbert}\tnote{**}\hspace{8pt}\tnote{1}& \textbf{Mistral-7B}\tnote{**} & \textbf{Llama2-13B}\tnote{**} & \textbf{Mixtral-8x7B}\tnote{**}& \textbf{Llama2-70B}\tnote{**} &\textbf{QWEN1.5-72B}\tnote{**}& \textbf{Ensemble}\tnote{**}\hspace{8pt}\tnote{2}& \textbf{GPT-3.5}\tnote{**} & \textbf{GPT-4} \\
\hline
Atelectasis & 0.837 & 0.978 & 0.861 & 0.765 & 0.778 & 0.978 & 0.974 & 0.982 & 0.896 & 0.978 \\ 
Cardiomegaly & 0.758 & 0.424 & 0.821 & 0.655 & 0.909 & 0.885 & 0.902 & 0.933 & 0.865 & 0.946 \\ 
Consolidation & 0.831 & 0.729 & 0.438 & 0.473 & 0.549 & 0.513 & 0.624 & 0.565 & 0.567 & 0.867 \\ 
Edema & 0.748 & 0.895 & 0.614 & 0.615 & 0.877 & 0.932 & 0.662 & 0.875 & 0.904 & 0.972 \\ 
Lung Lesion & 0.593 & 0.645 & 0.493 & 0.620 & 0.558 & 0.646 & 0.547 & 0.598 & 0.697 & 0.706 \\ 
Lung Opacity & 0.766 & 0.924 & 0.942 & 0.932 & 0.961 & 0.946 & 0.962 & 0.968 & 0.936 & 0.951 \\ 
Pleural Other & 0.725 & 0.889 & 0.532 & 0.475 & 0.743 & 0.587 & 0.674 & 0.694 & 0.653 & 0.865 \\ 
Pleural Effusion & 0.848 & 0.891 & 0.964 & 0.939 & 0.984 & 0.979 & 0.990 & 0.989 & 0.984 & 0.99 \\ 
Pneumonia & 0.474 & 0.870 & 0.880 & 0.868 & 0.818 & 0.917 & 0.875 & 0.892 & 0.921 & 0.923 \\ 
Pneumothorax & 0.464 & 0.773 & 0.973 & 0.947 & 1.0 & 0.947 & 1.0 & 1.0 & 0.947 & 1.0 \\ 
Support Devices & 0.856 & 0.570 & 0.879 & 0.764 & 0.902 & 0.884 & 0.902 & 0.907 & 0.846 & 0.891 \\ 
Enl. Cardiomediastinum & 0.222 & 0.370 & 0.472 & 0.687 & 0.952 & 0.902 & 0.899 & 0.952 & 0.909 & 0.975 \\ 
Fracture & 0.836 & 0.855 & 0.783 & 0.887 & 0.818 & 0.696 & 0.849 & 0.821 & 0.821 & 0.909 \\ 
\specialrule{1pt}{0pt}{0pt}
Micro F1 & 0.861 & 0.885 & 0.905 & 0.908 & 0.944 & 0.950 & 0.952 & 0.96 & 0.956 & 0.975 \\ 
Macro F1 & 0.689 & 0.755 & 0.743 & 0.741 & 0.835 & 0.832 & 0.835 & 0.86 & 0.842 & 0.921 \\ 
\specialrule{1pt}{0pt}{0pt}
\end{tabular}
}
\begin{tablenotes}
\scriptsize
\item[1] CheXpert-labeler and CheXbert do not take prompts, but are still listed here for reference 
\item[2] The Ensemble is constructed by combining the predictions of Mixtral-8x7B, Llama2-70B and QWEN1.5-72B through a majority vote. 
\item[*] There was a statistically significant difference in the performance compared to GPT-4 (p<0.05)
\item[**] There was a statistically significant difference in the performance compared to GPT-4 (p<0.01)

\end{tablenotes}
\end{threeparttable}
\label{tab:zero_shot_binary_institutional}
\end{table}

For the second experiment on the institutional dataset, the models were again prompted with a few-shot prompt. Table \ref{tab:few_shot_binary_institutional} shows the results for this task. The examples given to the model were chosen randomly, with examples chosen to include a positive label for each finding in as few reports as possible. This resulted in four example reports for this case. For the few-shot prompt, GPT-4 achieved the best performance with a micro F1-score of 0.973. 
QWEN1.5-72B, Llama2-70B, and Mixtral-8x7B achieved performances of 0.965, 0.965, and 0.963, respectively. Overall, all models except GPT-4 improved their performance with few-shot prompting. The improvement ranged from 5.1 percentage points for Mistral 7B to 1.3 percentage points for QWEN-1.5-72B. The ensemble model here showed very similar performance to GPT-4 with a Micro F-1 score of 0.971.

For both the zero-shot and few-shot prompts, all LLM models outperformed the CheXpert-labeler and CheXbert baselines on the institutional dataset, which achieved micro-F1 scores of 0.861 and 0.885, respectively. 

\begin{table}[ht]
\caption{F1-scores for few-shot Prompting on the Institutional Dataset for the binary case, with random few-shot examples}
\centering
\begin{threeparttable}
\resizebox{\textwidth}{!}{
\begin{tabular}{lcccccccccc}
\textbf{Finding} & \textbf{CheXpert}\tnote{**}\hspace{8pt}\tnote{1}& \textbf{CheXbert}\tnote{**}\hspace{8pt}\tnote{1}& \textbf{Mistral-7B}\tnote{**} & \textbf{Llama2-13B}\tnote{**} & \textbf{Mixtral-8x7B}\tnote{**}& \textbf{Llama2-70B}\tnote{**} &\textbf{QWEN1.5-72B}\tnote{**}& \textbf{Ensemble} \tnote{2}& \textbf{GPT-3.5} & \textbf{GPT-4} \\
\hline
Atelectasis & 0.837 & 0.978 & 0.974 & 0.970 & 0.945 & 0.982 & 0.982 & 0.978 & 0.974 & 0.975 \\ 
Cardiomegaly & 0.758 & 0.424 & 0.889 & 0.739 & 0.889 & 0.876 & 0.909 & 0.906 & 0.956 & 0.891 \\ 
Consolidation & 0.831 & 0.729 & 0.645 & 0.876 & 0.679 & 0.655 & 0.709 & 0.690 & 0.581 & 0.897 \\ 
Edema & 0.748 & 0.895 & 0.694 & 0.909 & 0.940 & 0.918 & 0.885 & 0.971 & 0.941 & 0.963 \\ 
Lung Lesion & 0.593 & 0.645 & 0.747 & 0.742 & 0.729 & 0.804 & 0.594 & 0.779 & 0.667 & 0.780 \\ 
Lung Opacity & 0.766 & 0.924 & 0.932 & 0.953 & 0.970 & 0.962 & 0.962 & 0.976 & 0.962 & 0.953 \\ 
Pleural Other & 0.725 & 0.889 & 0.661 & 0.342 & 0.696 & 0.673 & 0.731 & 0.697 & 0.695 & 0.821 \\ 
Pleural Effusion & 0.848 & 0.891 & 0.979 & 0.955 & 0.962 & 0.964 & 0.984 & 0.989 & 0.990 & 0.990 \\ 
Pneumonia & 0.474 & 0.870 & 0.876 & 0.855 & 0.880 & 0.915 & 0.869 & 0.929 & 0.906 & 0.933 \\ 
Pneumothorax & 0.464 & 0.773 & 0.919 & 0.919 & 0.944 & 0.947 & 1.0 & 0.973 & 0.973 & 1.0 \\ 
Support Devices & 0.856 & 0.57 & 0.879 & 0.667 & 0.875 & 0.884 & 0.915 & 0.887 & 0.880 & 0.853 \\ 
Enl. Cardiomediastinum & 0.222 & 0.370 & 0.882 & 0.776 & 0.896 & 0.919 & 0.959 & 0.950 & 0.974 & 0.943 \\ 
Fracture & 0.836 & 0.855 & 0.893 & 0.887 & 0.833 & 0.821 & 0.876 & 0.885 & 0.857 & 0.923 \\ 
\specialrule{1pt}{0pt}{0pt}
Micro F1 & 0.861 & 0.885 & 0.956 & 0.956 & 0.963 & 0.965 & 0.965 & 0.971 & 0.966 & 0.973 \\ 
Macro F1 & 0.689 & 0.755 & 0.844 & 0.815 & 0.864 & 0.871 & 0.875 & 0.893 & 0.874 & 0.917 \\ 
\specialrule{1pt}{0pt}{0pt}
\end{tabular}
}
\begin{tablenotes}
\scriptsize
\item[1] CheXpert-labeler and CheXbert do not take prompts, but are still listed here for reference 
\item[2] The Ensemble is constructed by combining the predictions of Mixtral-8x7B, Llama2-70B and QWEN1.5-72B through a majority vote. 
\item[*] There was a statistically significant difference in the performance compared to GPT-4 (p<0.05)
\item[**] There was a statistically significant difference in the performance compared to GPT-4 (p<0.01)
\end{tablenotes}

\end{threeparttable}
\label{tab:few_shot_binary_institutional}
\end{table}

Table \ref{tab:few_shot_multiclass_institutional} shows the Cohen's Kappa scores for the same few-shot prompts as above. For this analysis, the labels ``No'', ``Maybe'', and ``Yes'' were treated as 3 separate classes.

\begin{table}[ht]
\caption{Cohens Kappa Scores for few-shot Prompting on the Institutional Dataset for the multi-class case, with random few-shot examples.}
\centering
\begin{threeparttable}
\resizebox{\textwidth}{!}{
\begin{tabular}{lccccccccc}
\textbf{Finding} & \textbf{CheXpert}\tnote{1} & \textbf{CheXbert}\tnote{1} & \textbf{Mistral-7B} & \textbf{Llama2-13B} & \textbf{Mixtral-8x7B} & \textbf{Llama2-70B} &\textbf{QWEN1.5-72B} & \textbf{GPT-3.5} & \textbf{GPT-4} \\
\hline
Atelectasis & 0.759 & 0.925 & 0.891 & 0.876 & 0.833 & 0.839 & 0.916 & 0.892 & 0.849 \\ 
Cardiomegaly & 0.715 & 0.337 & 0.856 & 0.702 & 0.855 & 0.851 & 0.870 & 0.930 & 0.840 \\ 
Consolidation & 0.779 & 0.674 & 0.600 & 0.854 & 0.641 & 0.599 & 0.656 & 0.535 & 0.876 \\ 
Edema & 0.697 & 0.854 & 0.616 & 0.867 & 0.890 & 0.865 & 0.844 & 0.893 & 0.928 \\ 
Lung Lesion & 0.461 & 0.492 & 0.626 & 0.565 & 0.571 & 0.653 & 0.470 & 0.587 & 0.653 \\ 
Lung Opacity & 0.586 & 0.840 & 0.860 & 0.878 & 0.940 & 0.924 & 0.905 & 0.854 & 0.901 \\ 
Pleural Other & 0.702 & 0.806 & 0.535 & 0.260 & 0.564 & 0.566 & 0.618 & 0.599 & 0.714 \\ 
Pleural Effusion & 0.778 & 0.814 & 0.938 & 0.890 & 0.916 & 0.862 & 0.932 & 0.926 & 0.950 \\ 
Pneumonia & 0.315 & 0.787 & 0.789 & 0.772 & 0.815 & 0.704 & 0.788 & 0.795 & 0.863 \\ 
Pneumothorax & 0.433 & 0.739 & 0.916 & 0.916 & 0.942 & 0.918 & 0.973 & 0.944 & 0.973 \\ 
Support Devices & 0.801 & 0.308 & 0.834 & 0.591 & 0.825 & 0.840 & 0.882 & 0.840 & 0.790 \\ 
Enl. Cardiomediastinum & 0.012 & 0.239 & 0.840 & 0.730 & 0.854 & 0.889 & 0.925 & 0.933 & 0.899 \\ 
Fracture & 0.796 & 0.807 & 0.849 & 0.842 & 0.806 & 0.770 & 0.841 & 0.809 & 0.893 \\ 
\specialrule{1.5pt}{0pt}{0pt}
Average & 0.603 & 0.663 & 0.781 & 0.750 & 0.804 & 0.791 & 0.817 & 0.810 & 0.856 \\ 
\specialrule{1.5pt}{0pt}{0pt}
\end{tabular}
}
\begin{tablenotes}
\scriptsize
\item[1] CheXpert-labeler and CheXbert do not take prompts, but are still listed here for reference 
\end{tablenotes}
\end{threeparttable}
\label{tab:few_shot_multiclass_institutional}
\end{table}

\section{Discussion}\label{sec4}
In our study, we show that open-source generalist LLMs are able to consistently outperform the CheXbert model, which was specifically tuned for the task of radiology report classification. Furthermore, they come very close to the performance of GPT-4, which is a much larger model. While it is unknown how many parameters GPT-4 exactly has, its predecessor GPT-3 had a total of 175B parameters \cite{brown_language_2020}. This highlights how well the small models Mistral-7B and Llama13B performed on this task, given that they have an order of magnitude fewer parameters. When examining their performance on the institutional datasets, these small models are also the ones that had the largest performance increase through the use of few-shot prompting. 

Model performance also varied significantly between the labels on both datasets. For example, while QWEN1.5-72B, Mistral-7B, and Mixtral-8x7B performed well overall on the zero-shot task for the ImaGenome dataset, they show very poor performance on the pneumonia label. Examination of the raw model outputs before preprocessing shows that the models did not stick to the choices given in the prompt, but instead also produced results such as ``Possible'', ``Uncertain'', ``Suspect'', and ``Possibly'', which could be classified as ``Maybe''. When the model performance was calculated, all of these labels that did not adhere to the template in the prompt were converted to ``No'', which was necessary to obtain an unambiguous and computer-readable output. QWEN1.5-72B used the output label ``Possible'' 30 times for the label ``Pneumonia''. This illustrates a potential problem with using generative models for classification tasks. While it is possible to provide classification labels for the LLM in the prompt, there is no guarantee that the model output will actually be limited to the options provided in the prompt. For the institutional dataset, there are three particular classes, ``Consolidation'', ``Lung Lesion'', and ``Pleural Other'', that showed poor performance across the range of models included in this study. There may be several factors that contribute to these classes being particularly difficult for the models to classify. It is possible that these classes are not defined or explained clearly enough in the prompt for the models to pick up these findings, especially because they are findings that, when present in the report, are often not referred to with the exact same word as the label that we are assigning. The models performed particularly well on the findings ``Pneumothorax'' and ``Atelectasis''. These are the findings that are most often mentioned literally in the report. Optimizing the performance of the LLMs across all findings could potentially be achieved by creating longer and more explicit prompts and should be the subject of future work.

In addition, we were able to improve the overall performance of the open-source models on the institutional dataset by combining the three models with the best individual performance (Mixtral-8x7B, Llama2-70B, QWEN1.5-72B) into an ensemble by taking a majority vote among the three models for each individual report and finding. This ensemble was particularly capable because its individual models had different labels on which they performed well. This ensemble of open-source models was able to very closely match the performance of GPT-4 on the institutional dataset with few-shot prompts, with micro F1-scores of 0.971 and 0.973, respectively. The ensemble achieved higher micro and macro F1-Scores than GPT-3.5 Turbo on the zero-shot task for the ImaGenome dataset and both tasks for the institutional dataset.

These results show that open source LLMs can serve as a viable alternative to GPT-4, as they are close in performance and offer several other significant advantages. First, there is a significant cost advantage. As long as sufficient computing resources are available, there is no additional cost to classify reports using open source models, whereas the GPT API is charged per 1000 tokens. This can be very costly when applied to the tens of thousands of reports that would be required to create a dataset. In the context of providing labels for downstream artificial intelligence tasks, the availability of sufficient computational resources can be assumed, as the training of these models itself is also computationally expensive. On top of this, many academic medical centers have access to GPU-enabled compute clusters that could be used to run these models. In addition, model quantization is a powerful technique to reduce the size of a model by reducing the numerical precision. In its Int4 quantized version, the Llama-2 70B model can run on 48 GB of VRAM, allowing it to run on a single graphics card. Even without the local computing capabilities to run the models, they still have a price advantage over GPT-4 when used through external cloud providers, as they are more cost-effective to run. 

In addition, the use of open source models ensures consistency and reproducibility over time because the models are local and therefore not subject to updates in their model weights or the associated cloud infrastructure and API. Furthermore, open source models have the advantage of outputting the token prediction probabilities, which can be further used to determine the uncertainty of the predictions. 

Another key advantage is privacy and compliance with associated regulatory requirements. By using open source LLMs that run locally, privacy can be ensured because the data does not have to leave the hospital network to be processed on a remote server, as is the case with a proprietary model. This is especially important when dealing with sensitive medical data, such as patient information contained in radiology reports.  

Our study also has some limitations. First, the prompts were all written and tested for the first time on GPT-4. Some modifications were made later to make the same prompts work for all models, but this may still give an advantage to GPT-4. Also, it is unclear and difficult to judge whether the prompts we used for the experiments are the ideal prompts for this scenario. We tried different prompts on a development set of 100 reports, but did not do an exhaustive search. In addition, both datasets had a class imbalance, with the Imagenome dataset having only 10 positive samples for ``Pneumothorax''. We mitigated the effect by oversampling for rare findings in the institutional dataset, but still observed a significant class imbalance. While this reflects a clinical reality - some findings are rarer than others - it still has statistical implications. 

Overall, our results demonstrate that open-source LLMs are a viable and valuable alternative to proprietary models for medical tasks such as radiology report classification.

\FloatBarrier

\section*{Acknowledgements}
Much of the computation resources required for this research was performed on computational hardware generously provided by the Massachusetts Life Sciences Center (https://www.masslifesciences.com/).
Access to GPT-3.5 and GPT-4 was through a HIPAA-compliant cloud implementation.

\bibliographystyle{unsrtnat}
\bibliography{references.bib}

\begin{thebibliography}{26}
\providecommand{\natexlab}[1]{#1}
\providecommand{\url}[1]{\texttt{#1}}
\expandafter\ifx\csname urlstyle\endcsname\relax
  \providecommand{\doi}[1]{doi: #1}\else
  \providecommand{\doi}{doi: \begingroup \urlstyle{rm}\Url}\fi

\bibitem[Thirunavukarasu et~al.(2023)Thirunavukarasu, Ting, Elangovan, Gutierrez, Tan, and Ting]{thirunavukarasu_large_2023}
Arun~James Thirunavukarasu, Darren Shu~Jeng Ting, Kabilan Elangovan, Laura Gutierrez, Ting~Fang Tan, and Daniel Shu~Wei Ting.
\newblock Large language models in medicine.
\newblock \emph{Nature Medicine}, 29\penalty0 (8):\penalty0 1930--1940, August 2023.
\newblock ISSN 1546-170X.
\newblock \doi{10.1038/s41591-023-02448-8}.
\newblock URL \url{https://www.nature.com/articles/s41591-023-02448-8}.
\newblock Number: 8 Publisher: Nature Publishing Group.

\bibitem[Zhao et~al.(2023)Zhao, Zhou, Li, Tang, Wang, Hou, Min, Zhang, Zhang, Dong, Du, Yang, Chen, Chen, Jiang, Ren, Li, Tang, Liu, Liu, Nie, and Wen]{zhao_survey_2023}
Wayne~Xin Zhao, Kun Zhou, Junyi Li, Tianyi Tang, Xiaolei Wang, Yupeng Hou, Yingqian Min, Beichen Zhang, Junjie Zhang, Zican Dong, Yifan Du, Chen Yang, Yushuo Chen, Zhipeng Chen, Jinhao Jiang, Ruiyang Ren, Yifan Li, Xinyu Tang, Zikang Liu, Peiyu Liu, Jian-Yun Nie, and Ji-Rong Wen.
\newblock A {Survey} of {Large} {Language} {Models}, November 2023.
\newblock URL \url{http://arxiv.org/abs/2303.18223}.
\newblock arXiv:2303.18223 [cs].

\bibitem[Adams et~al.(2023)Adams, Truhn, Busch, Kader, Niehues, Makowski, and Bressem]{adams_leveraging_2023}
Lisa~C. Adams, Daniel Truhn, Felix Busch, Avan Kader, Stefan~M. Niehues, Marcus~R. Makowski, and Keno~K. Bressem.
\newblock Leveraging {GPT}-4 for {Post} {Hoc} {Transformation} of {Free}-{Text} {Radiology} {Reports} into {Structured} {Reporting}: {A} {Multilingual} {Feasibility} {Study}.
\newblock \emph{Radiology}, page 230725, April 2023.
\newblock ISSN 0033-8419.
\newblock \doi{10.1148/radiol.230725}.
\newblock URL \url{https://pubs.rsna.org/doi/10.1148/radiol.230725}.

\bibitem[Liu et~al.(2023)Liu, Hyland, Bannur, Bouzid, Castro, Wetscherek, Tinn, Sharma, Pérez-García, Schwaighofer, Rajpurkar, Khanna, Poon, Usuyama, Thieme, Nori, Lungren, Oktay, and Alvarez-Valle]{liu_exploring_2023}
Qianchu Liu, Stephanie Hyland, Shruthi Bannur, Kenza Bouzid, Daniel~C. Castro, Maria~Teodora Wetscherek, Robert Tinn, Harshita Sharma, Fernando Pérez-García, Anton Schwaighofer, Pranav Rajpurkar, Sameer~Tajdin Khanna, Hoifung Poon, Naoto Usuyama, Anja Thieme, Aditya~V. Nori, Matthew~P. Lungren, Ozan Oktay, and Javier Alvarez-Valle.
\newblock Exploring the {Boundaries} of {GPT}-4 in {Radiology}, October 2023.
\newblock URL \url{http://arxiv.org/abs/2310.14573}.
\newblock arXiv:2310.14573 [cs].

\bibitem[Infante et~al.(2024)Infante, Gaudino, Orsini, Del~Ciello, Gullì, Merlino, Natale, Iezzi, and Sala]{infante_large_2024}
A.~Infante, S.~Gaudino, F.~Orsini, A.~Del~Ciello, C.~Gullì, B.~Merlino, L.~Natale, R.~Iezzi, and E.~Sala.
\newblock Large language models ({LLMs}) in the evaluation of emergency radiology reports: performance of {ChatGPT}-4, {Perplexity}, and {Bard}.
\newblock \emph{Clinical Radiology}, 79\penalty0 (2):\penalty0 102--106, February 2024.
\newblock ISSN 0009-9260.
\newblock \doi{10.1016/j.crad.2023.11.011}.
\newblock URL \url{https://www.sciencedirect.com/science/article/pii/S0009926023005342}.

\bibitem[Wu et~al.(2021)Wu, Agu, Lourentzou, Sharma, Paguio, Yao, Dee, Mitchell, Kashyap, Giovannini, Celi, Syeda-Mahmood, and Moradi]{wu_chest_2021}
J.~Wu, N.~Agu, I.~Lourentzou, A.~Sharma, J.~Paguio, J.~S. Yao, E.~C. Dee, W.~Mitchell, S.~Kashyap, A.~Giovannini, L.~A. Celi, T.~Syeda-Mahmood, and M.~Moradi.
\newblock Chest {ImaGenome} {Dataset}, 2021.
\newblock URL \url{https://doi.org/10.13026/wv01-y230}.
\newblock Version Number: 1.0.0 Published: PhysioNet.

\bibitem[Goldberger et~al.(2000)Goldberger, Amaral, Glass, Hausdorff, Ivanov, Mark, and Stanley]{goldberger_physiobank_2000}
A.~L. Goldberger, L.~Amaral, L.~Glass, J.~M. Hausdorff, P.~C. Ivanov, R.~G. Mark, and H.~E. Stanley.
\newblock {PhysioBank}, {PhysioToolkit}, and {PhysioNet}: {Components} of a new research resource for complex physiologic signals.
\newblock \emph{Circulation}, 101\penalty0 (23):\penalty0 e215--e220, 2000.
\newblock Publisher: American Heart Association.

\bibitem[Johnson et~al.(2019{\natexlab{a}})Johnson, Pollard, Berkowitz, and {others}]{johnson_mimic-cxr_2019}
A.E.W. Johnson, T.J. Pollard, S.J. Berkowitz, and {others}.
\newblock {MIMIC}-{CXR}, a de-identified publicly available database of chest radiographs with free-text reports.
\newblock \emph{Sci Data}, 6:\penalty0 317, 2019{\natexlab{a}}.
\newblock \doi{10.1038/s41597-019-0322-0}.
\newblock URL \url{https://doi.org/10.1038/s41597-019-0322-0}.
\newblock Publisher: Nature Publishing Group.

\bibitem[Johnson et~al.(2019{\natexlab{b}})Johnson, Pollard, Mark, Berkowitz, and Horng]{johnson_mimic-cxr_2019-1}
A.~Johnson, T.~Pollard, R.~Mark, S.~Berkowitz, and S.~Horng.
\newblock {MIMIC}-{CXR} {Database}, 2019{\natexlab{b}}.
\newblock URL \url{https://doi.org/10.13026/C2JT1Q}.
\newblock Version Number: 2.0.0.

\bibitem[Smit et~al.(2020)Smit, Jain, Rajpurkar, Pareek, Ng, and Lungren]{smit_CheXbert_2020}
Akshay Smit, Saahil Jain, Pranav Rajpurkar, Anuj Pareek, Andrew~Y. Ng, and Matthew~P. Lungren.
\newblock {CheXbert}: {Combining} {Automatic} {Labelers} and {Expert} {Annotations} for {Accurate} {Radiology} {Report} {Labeling} {Using} {BERT}, October 2020.
\newblock URL \url{http://arxiv.org/abs/2004.09167}.
\newblock arXiv:2004.09167 [cs].

\bibitem[Irvin et~al.(2019)Irvin, Rajpurkar, Ko, Yu, Ciurea-Ilcus, Chute, Marklund, Haghgoo, Ball, Shpanskaya, Seekins, Mong, Halabi, Sandberg, Jones, Larson, Langlotz, Patel, Lungren, and Ng]{irvin_chexpert_2019}
Jeremy Irvin, Pranav Rajpurkar, Michael Ko, Yifan Yu, Silviana Ciurea-Ilcus, Chris Chute, Henrik Marklund, Behzad Haghgoo, Robyn Ball, Katie Shpanskaya, Jayne Seekins, David~A. Mong, Safwan~S. Halabi, Jesse~K. Sandberg, Ricky Jones, David~B. Larson, Curtis~P. Langlotz, Bhavik~N. Patel, Matthew~P. Lungren, and Andrew~Y. Ng.
\newblock {CheXpert}: {A} {Large} {Chest} {Radiograph} {Dataset} with {Uncertainty} {Labels} and {Expert} {Comparison}, January 2019.
\newblock URL \url{http://arxiv.org/abs/1901.07031}.
\newblock arXiv:1901.07031 [cs, eess].

\bibitem[Kwon et~al.(2023)Kwon, Li, Zhuang, Sheng, Zheng, Yu, Gonzalez, Zhang, and Stoica]{kwon_efficient_2023}
Woosuk Kwon, Zhuohan Li, Siyuan Zhuang, Ying Sheng, Lianmin Zheng, Cody~Hao Yu, Joseph~E. Gonzalez, Hao Zhang, and Ion Stoica.
\newblock Efficient {Memory} {Management} for {Large} {Language} {Model} {Serving} with {PagedAttention}, September 2023.
\newblock URL \url{http://arxiv.org/abs/2309.06180}.
\newblock arXiv:2309.06180 [cs].

\bibitem[Lin et~al.(2023)Lin, Tang, Tang, Yang, Dang, Gan, and Han]{lin_awq_2023}
Ji~Lin, Jiaming Tang, Haotian Tang, Shang Yang, Xingyu Dang, Chuang Gan, and Song Han.
\newblock {AWQ}: {Activation}-aware {Weight} {Quantization} for {LLM} {Compression} and {Acceleration}, October 2023.
\newblock URL \url{http://arxiv.org/abs/2306.00978}.
\newblock arXiv:2306.00978 [cs].

\bibitem[Brown et~al.(2020)Brown, Mann, Ryder, Subbiah, Kaplan, Dhariwal, Neelakantan, Shyam, Sastry, Askell, Agarwal, Herbert-Voss, Krueger, Henighan, Child, Ramesh, Ziegler, Wu, Winter, Hesse, Chen, Sigler, Litwin, Gray, Chess, Clark, Berner, McCandlish, Radford, Sutskever, and Amodei]{brown_language_2020}
Tom~B. Brown, Benjamin Mann, Nick Ryder, Melanie Subbiah, Jared Kaplan, Prafulla Dhariwal, Arvind Neelakantan, Pranav Shyam, Girish Sastry, Amanda Askell, Sandhini Agarwal, Ariel Herbert-Voss, Gretchen Krueger, Tom Henighan, Rewon Child, Aditya Ramesh, Daniel~M. Ziegler, Jeffrey Wu, Clemens Winter, Christopher Hesse, Mark Chen, Eric Sigler, Mateusz Litwin, Scott Gray, Benjamin Chess, Jack Clark, Christopher Berner, Sam McCandlish, Alec Radford, Ilya Sutskever, and Dario Amodei.
\newblock Language {Models} are {Few}-{Shot} {Learners}, July 2020.
\newblock URL \url{http://arxiv.org/abs/2005.14165}.
\newblock arXiv:2005.14165 [cs].

\bibitem[Touvron et~al.(2023)Touvron, Martin, Stone, Albert, Almahairi, Babaei, Bashlykov, Batra, Bhargava, Bhosale, Bikel, Blecher, Ferrer, Chen, Cucurull, Esiobu, Fernandes, Fu, Fu, Fuller, Gao, Goswami, Goyal, Hartshorn, Hosseini, Hou, Inan, Kardas, Kerkez, Khabsa, Kloumann, Korenev, Koura, Lachaux, Lavril, Lee, Liskovich, Lu, Mao, Martinet, Mihaylov, Mishra, Molybog, Nie, Poulton, Reizenstein, Rungta, Saladi, Schelten, Silva, Smith, Subramanian, Tan, Tang, Taylor, Williams, Kuan, Xu, Yan, Zarov, Zhang, Fan, Kambadur, Narang, Rodriguez, Stojnic, Edunov, and Scialom]{touvron_llama_2023}
Hugo Touvron, Louis Martin, Kevin Stone, Peter Albert, Amjad Almahairi, Yasmine Babaei, Nikolay Bashlykov, Soumya Batra, Prajjwal Bhargava, Shruti Bhosale, Dan Bikel, Lukas Blecher, Cristian~Canton Ferrer, Moya Chen, Guillem Cucurull, David Esiobu, Jude Fernandes, Jeremy Fu, Wenyin Fu, Brian Fuller, Cynthia Gao, Vedanuj Goswami, Naman Goyal, Anthony Hartshorn, Saghar Hosseini, Rui Hou, Hakan Inan, Marcin Kardas, Viktor Kerkez, Madian Khabsa, Isabel Kloumann, Artem Korenev, Punit~Singh Koura, Marie-Anne Lachaux, Thibaut Lavril, Jenya Lee, Diana Liskovich, Yinghai Lu, Yuning Mao, Xavier Martinet, Todor Mihaylov, Pushkar Mishra, Igor Molybog, Yixin Nie, Andrew Poulton, Jeremy Reizenstein, Rashi Rungta, Kalyan Saladi, Alan Schelten, Ruan Silva, Eric~Michael Smith, Ranjan Subramanian, Xiaoqing~Ellen Tan, Binh Tang, Ross Taylor, Adina Williams, Jian~Xiang Kuan, Puxin Xu, Zheng Yan, Iliyan Zarov, Yuchen Zhang, Angela Fan, Melanie Kambadur, Sharan Narang, Aurelien Rodriguez, Robert Stojnic, Sergey Edunov, and Thomas
  Scialom.
\newblock Llama 2: {Open} {Foundation} and {Fine}-{Tuned} {Chat} {Models}, July 2023.
\newblock URL \url{http://arxiv.org/abs/2307.09288}.
\newblock arXiv:2307.09288 [cs].

\bibitem[Jiang et~al.(2023)Jiang, Sablayrolles, Mensch, Bamford, Chaplot, Casas, Bressand, Lengyel, Lample, Saulnier, Lavaud, Lachaux, Stock, Scao, Lavril, Wang, Lacroix, and Sayed]{jiang_mistral_2023}
Albert~Q. Jiang, Alexandre Sablayrolles, Arthur Mensch, Chris Bamford, Devendra~Singh Chaplot, Diego de~las Casas, Florian Bressand, Gianna Lengyel, Guillaume Lample, Lucile Saulnier, Lélio~Renard Lavaud, Marie-Anne Lachaux, Pierre Stock, Teven~Le Scao, Thibaut Lavril, Thomas Wang, Timothée Lacroix, and William~El Sayed.
\newblock Mistral {7B}, October 2023.
\newblock URL \url{http://arxiv.org/abs/2310.06825}.
\newblock arXiv:2310.06825 [cs].

\bibitem[Jiang et~al.(2024)Jiang, Sablayrolles, Roux, Mensch, Savary, Bamford, Chaplot, Casas, Hanna, Bressand, Lengyel, Bour, Lample, Lavaud, Saulnier, Lachaux, Stock, Subramanian, Yang, Antoniak, Scao, Gervet, Lavril, Wang, Lacroix, and Sayed]{jiang_mixtral_2024}
Albert~Q. Jiang, Alexandre Sablayrolles, Antoine Roux, Arthur Mensch, Blanche Savary, Chris Bamford, Devendra~Singh Chaplot, Diego de~las Casas, Emma~Bou Hanna, Florian Bressand, Gianna Lengyel, Guillaume Bour, Guillaume Lample, Lélio~Renard Lavaud, Lucile Saulnier, Marie-Anne Lachaux, Pierre Stock, Sandeep Subramanian, Sophia Yang, Szymon Antoniak, Teven~Le Scao, Théophile Gervet, Thibaut Lavril, Thomas Wang, Timothée Lacroix, and William~El Sayed.
\newblock Mixtral of {Experts}, January 2024.
\newblock URL \url{http://arxiv.org/abs/2401.04088}.
\newblock arXiv:2401.04088 [cs].

\bibitem[Bai et~al.(2023)Bai, Bai, Chu, Cui, Dang, Deng, Fan, Ge, Han, Huang, Hui, Ji, Li, Lin, Lin, Liu, Liu, Lu, Lu, Ma, Men, Ren, Ren, Tan, Tan, Tu, Wang, Wang, Wang, Wu, Xu, Xu, Yang, Yang, Yang, Yang, Yao, Yu, Yuan, Yuan, Zhang, Zhang, Zhang, Zhang, Zhou, Zhou, Zhou, and Zhu]{bai_qwen_2023}
Jinze Bai, Shuai Bai, Yunfei Chu, Zeyu Cui, Kai Dang, Xiaodong Deng, Yang Fan, Wenbin Ge, Yu~Han, Fei Huang, Binyuan Hui, Luo Ji, Mei Li, Junyang Lin, Runji Lin, Dayiheng Liu, Gao Liu, Chengqiang Lu, Keming Lu, Jianxin Ma, Rui Men, Xingzhang Ren, Xuancheng Ren, Chuanqi Tan, Sinan Tan, Jianhong Tu, Peng Wang, Shijie Wang, Wei Wang, Shengguang Wu, Benfeng Xu, Jin Xu, An~Yang, Hao Yang, Jian Yang, Shusheng Yang, Yang Yao, Bowen Yu, Hongyi Yuan, Zheng Yuan, Jianwei Zhang, Xingxuan Zhang, Yichang Zhang, Zhenru Zhang, Chang Zhou, Jingren Zhou, Xiaohuan Zhou, and Tianhang Zhu.
\newblock Qwen {Technical} {Report}, September 2023.
\newblock URL \url{http://arxiv.org/abs/2309.16609}.
\newblock arXiv:2309.16609 [cs].

\bibitem[OpenAI et~al.(2023)OpenAI, Achiam, Adler, Agarwal, Ahmad, Akkaya, Aleman, Almeida, Altenschmidt, Altman, Anadkat, Avila, Babuschkin, Balaji, Balcom, Baltescu, Bao, Bavarian, Belgum, Bello, Berdine, Bernadett-Shapiro, Berner, Bogdonoff, Boiko, Boyd, Brakman, Brockman, Brooks, Brundage, Button, Cai, Campbell, Cann, Carey, Carlson, Carmichael, Chan, Chang, Chantzis, Chen, Chen, Chen, Chen, Chen, Chess, Cho, Chu, Chung, Cummings, Currier, Dai, Decareaux, Degry, Deutsch, Deville, Dhar, Dohan, Dowling, Dunning, Ecoffet, Eleti, Eloundou, Farhi, Fedus, Felix, Fishman, Forte, Fulford, Gao, Georges, Gibson, Goel, Gogineni, Goh, Gontijo-Lopes, Gordon, Grafstein, Gray, Greene, Gross, Gu, Guo, Hallacy, Han, Harris, He, Heaton, Heidecke, Hesse, Hickey, Hickey, Hoeschele, Houghton, Hsu, Hu, Hu, Huizinga, Jain, Jain, Jang, Jiang, Jiang, Jin, Jin, Jomoto, Jonn, Jun, Kaftan, Kaiser, Kamali, Kanitscheider, Keskar, Khan, Kilpatrick, Kim, Kim, Kim, Kirchner, Kiros, Knight, Kokotajlo, Kondraciuk, Kondrich, Konstantinidis,
  Kosic, Krueger, Kuo, Lampe, Lan, Lee, Leike, Leung, Levy, Li, Lim, Lin, Lin, Litwin, Lopez, Lowe, Lue, Makanju, Malfacini, Manning, Markov, Markovski, Martin, Mayer, Mayne, McGrew, McKinney, McLeavey, McMillan, McNeil, Medina, Mehta, Menick, Metz, Mishchenko, Mishkin, Monaco, Morikawa, Mossing, Mu, Murati, Murk, Mély, Nair, Nakano, Nayak, Neelakantan, Ngo, Noh, Ouyang, O'Keefe, Pachocki, Paino, Palermo, Pantuliano, Parascandolo, Parish, Parparita, Passos, Pavlov, Peng, Perelman, Peres, Petrov, Pinto, Michael, Pokorny, Pokrass, Pong, Powell, Power, Power, Proehl, Puri, Radford, Rae, Ramesh, Raymond, Real, Rimbach, Ross, Rotsted, Roussez, Ryder, Saltarelli, Sanders, Santurkar, Sastry, Schmidt, Schnurr, Schulman, Selsam, Sheppard, Sherbakov, Shieh, Shoker, Shyam, Sidor, Sigler, Simens, Sitkin, Slama, Sohl, Sokolowsky, Song, Staudacher, Such, Summers, Sutskever, Tang, Tezak, Thompson, Tillet, Tootoonchian, Tseng, Tuggle, Turley, Tworek, Uribe, Vallone, Vijayvergiya, Voss, Wainwright, Wang, Wang, Wang, Ward,
  Wei, Weinmann, Welihinda, Welinder, Weng, Weng, Wiethoff, Willner, Winter, Wolrich, Wong, Workman, Wu, Wu, Wu, Xiao, Xu, Yoo, Yu, Yuan, Zaremba, Zellers, Zhang, Zhang, Zhao, Zheng, Zhuang, Zhuk, and Zoph]{openai_gpt-4_2023}
OpenAI, Josh Achiam, Steven Adler, Sandhini Agarwal, Lama Ahmad, Ilge Akkaya, Florencia~Leoni Aleman, Diogo Almeida, Janko Altenschmidt, Sam Altman, Shyamal Anadkat, Red Avila, Igor Babuschkin, Suchir Balaji, Valerie Balcom, Paul Baltescu, Haiming Bao, Mo~Bavarian, Jeff Belgum, Irwan Bello, Jake Berdine, Gabriel Bernadett-Shapiro, Christopher Berner, Lenny Bogdonoff, Oleg Boiko, Madelaine Boyd, Anna-Luisa Brakman, Greg Brockman, Tim Brooks, Miles Brundage, Kevin Button, Trevor Cai, Rosie Campbell, Andrew Cann, Brittany Carey, Chelsea Carlson, Rory Carmichael, Brooke Chan, Che Chang, Fotis Chantzis, Derek Chen, Sully Chen, Ruby Chen, Jason Chen, Mark Chen, Ben Chess, Chester Cho, Casey Chu, Hyung~Won Chung, Dave Cummings, Jeremiah Currier, Yunxing Dai, Cory Decareaux, Thomas Degry, Noah Deutsch, Damien Deville, Arka Dhar, David Dohan, Steve Dowling, Sheila Dunning, Adrien Ecoffet, Atty Eleti, Tyna Eloundou, David Farhi, Liam Fedus, Niko Felix, Simón~Posada Fishman, Juston Forte, Isabella Fulford, Leo Gao,
  Elie Georges, Christian Gibson, Vik Goel, Tarun Gogineni, Gabriel Goh, Rapha Gontijo-Lopes, Jonathan Gordon, Morgan Grafstein, Scott Gray, Ryan Greene, Joshua Gross, Shixiang~Shane Gu, Yufei Guo, Chris Hallacy, Jesse Han, Jeff Harris, Yuchen He, Mike Heaton, Johannes Heidecke, Chris Hesse, Alan Hickey, Wade Hickey, Peter Hoeschele, Brandon Houghton, Kenny Hsu, Shengli Hu, Xin Hu, Joost Huizinga, Shantanu Jain, Shawn Jain, Joanne Jang, Angela Jiang, Roger Jiang, Haozhun Jin, Denny Jin, Shino Jomoto, Billie Jonn, Heewoo Jun, Tomer Kaftan, Łukasz Kaiser, Ali Kamali, Ingmar Kanitscheider, Nitish~Shirish Keskar, Tabarak Khan, Logan Kilpatrick, Jong~Wook Kim, Christina Kim, Yongjik Kim, Hendrik Kirchner, Jamie Kiros, Matt Knight, Daniel Kokotajlo, Łukasz Kondraciuk, Andrew Kondrich, Aris Konstantinidis, Kyle Kosic, Gretchen Krueger, Vishal Kuo, Michael Lampe, Ikai Lan, Teddy Lee, Jan Leike, Jade Leung, Daniel Levy, Chak~Ming Li, Rachel Lim, Molly Lin, Stephanie Lin, Mateusz Litwin, Theresa Lopez, Ryan Lowe,
  Patricia Lue, Anna Makanju, Kim Malfacini, Sam Manning, Todor Markov, Yaniv Markovski, Bianca Martin, Katie Mayer, Andrew Mayne, Bob McGrew, Scott~Mayer McKinney, Christine McLeavey, Paul McMillan, Jake McNeil, David Medina, Aalok Mehta, Jacob Menick, Luke Metz, Andrey Mishchenko, Pamela Mishkin, Vinnie Monaco, Evan Morikawa, Daniel Mossing, Tong Mu, Mira Murati, Oleg Murk, David Mély, Ashvin Nair, Reiichiro Nakano, Rajeev Nayak, Arvind Neelakantan, Richard Ngo, Hyeonwoo Noh, Long Ouyang, Cullen O'Keefe, Jakub Pachocki, Alex Paino, Joe Palermo, Ashley Pantuliano, Giambattista Parascandolo, Joel Parish, Emy Parparita, Alex Passos, Mikhail Pavlov, Andrew Peng, Adam Perelman, Filipe de Avila~Belbute Peres, Michael Petrov, Henrique Ponde de~Oliveira Pinto, Michael, Pokorny, Michelle Pokrass, Vitchyr Pong, Tolly Powell, Alethea Power, Boris Power, Elizabeth Proehl, Raul Puri, Alec Radford, Jack Rae, Aditya Ramesh, Cameron Raymond, Francis Real, Kendra Rimbach, Carl Ross, Bob Rotsted, Henri Roussez, Nick Ryder,
  Mario Saltarelli, Ted Sanders, Shibani Santurkar, Girish Sastry, Heather Schmidt, David Schnurr, John Schulman, Daniel Selsam, Kyla Sheppard, Toki Sherbakov, Jessica Shieh, Sarah Shoker, Pranav Shyam, Szymon Sidor, Eric Sigler, Maddie Simens, Jordan Sitkin, Katarina Slama, Ian Sohl, Benjamin Sokolowsky, Yang Song, Natalie Staudacher, Felipe~Petroski Such, Natalie Summers, Ilya Sutskever, Jie Tang, Nikolas Tezak, Madeleine Thompson, Phil Tillet, Amin Tootoonchian, Elizabeth Tseng, Preston Tuggle, Nick Turley, Jerry Tworek, Juan Felipe~Cerón Uribe, Andrea Vallone, Arun Vijayvergiya, Chelsea Voss, Carroll Wainwright, Justin~Jay Wang, Alvin Wang, Ben Wang, Jonathan Ward, Jason Wei, C.~J. Weinmann, Akila Welihinda, Peter Welinder, Jiayi Weng, Lilian Weng, Matt Wiethoff, Dave Willner, Clemens Winter, Samuel Wolrich, Hannah Wong, Lauren Workman, Sherwin Wu, Jeff Wu, Michael Wu, Kai Xiao, Tao Xu, Sarah Yoo, Kevin Yu, Qiming Yuan, Wojciech Zaremba, Rowan Zellers, Chong Zhang, Marvin Zhang, Shengjia Zhao, Tianhao
  Zheng, Juntang Zhuang, William Zhuk, and Barret Zoph.
\newblock {GPT}-4 {Technical} {Report}, December 2023.
\newblock URL \url{http://arxiv.org/abs/2303.08774}.
\newblock arXiv:2303.08774 [cs].

\bibitem[Harris et~al.(2020)Harris, Millman, Walt, Gommers, Virtanen, Cournapeau, Wieser, Taylor, Berg, Smith, Kern, Picus, Hoyer, Kerkwijk, Brett, Haldane, Río, Wiebe, Peterson, Gérard-Marchant, Sheppard, Reddy, Weckesser, Abbasi, Gohlke, and Oliphant]{harris_array_2020}
Charles~R. Harris, K.~Jarrod Millman, Stéfan J. van~der Walt, Ralf Gommers, Pauli Virtanen, David Cournapeau, Eric Wieser, Julian Taylor, Sebastian Berg, Nathaniel~J. Smith, Robert Kern, Matti Picus, Stephan Hoyer, Marten H.~van Kerkwijk, Matthew Brett, Allan Haldane, Jaime Fernández~del Río, Mark Wiebe, Pearu Peterson, Pierre Gérard-Marchant, Kevin Sheppard, Tyler Reddy, Warren Weckesser, Hameer Abbasi, Christoph Gohlke, and Travis~E. Oliphant.
\newblock Array programming with {NumPy}.
\newblock \emph{Nature}, 585\penalty0 (7825):\penalty0 357--362, September 2020.
\newblock \doi{10.1038/s41586-020-2649-2}.
\newblock URL \url{https://doi.org/10.1038/s41586-020-2649-2}.
\newblock Publisher: Springer Science and Business Media LLC.

\bibitem[McKinney(2010)]{mckinney_data_2010}
Wes McKinney.
\newblock Data {Structures} for {Statistical} {Computing} in {Python}.
\newblock In Stéfan van~der Walt and Jarrod Millman, editors, \emph{Proceedings of the 9th {Python} in {Science} {Conference}}, pages 56 -- 61, 2010.
\newblock \doi{10.25080/Majora-92bf1922-00a}.

\bibitem[Pedregosa et~al.(2011)Pedregosa, Varoquaux, Gramfort, Michel, Thirion, Grisel, Blondel, Prettenhofer, Weiss, Dubourg, Vanderplas, Passos, Cournapeau, Brucher, Perrot, and Duchesnay]{pedregosa_scikit-learn_2011}
Fabian Pedregosa, Gaël Varoquaux, Alexandre Gramfort, Vincent Michel, Bertrand Thirion, Olivier Grisel, Mathieu Blondel, Peter Prettenhofer, Ron Weiss, Vincent Dubourg, Jake Vanderplas, Alexandre Passos, David Cournapeau, Matthieu Brucher, Matthieu Perrot, and Édouard Duchesnay.
\newblock Scikit-learn: {Machine} {Learning} in {Python}.
\newblock \emph{Journal of Machine Learning Research}, 12\penalty0 (85):\penalty0 2825--2830, 2011.
\newblock URL \url{http://jmlr.org/papers/v12/pedregosa11a.html}.

\bibitem[Hunter(2023)]{hunter_matplotlib_2023}
J.~D. Hunter.
\newblock Matplotlib: {A} {2D} graphics environment.
\newblock \emph{Computing in Science \& Engineering}, 9\penalty0 (3):\penalty0 90--95, April 2023.
\newblock \doi{10.1109/MCSE.2007.55}.
\newblock URL \url{https://github.com/matplotlib/matplotlib/blob/31b374ced4604e45b77c284801899a2f0cd158fa/CITATION.bib}.
\newblock original-date: 2011-02-19T03:17:12Z.

\bibitem[Seabold and Perktold(2010)]{seabold_statsmodels_2010}
Skipper Seabold and Josef Perktold.
\newblock statsmodels: {Econometric} and statistical modeling with python.
\newblock In \emph{9th {Python} in {Science} {Conference}}, 2010.

\bibitem[Waskom(2021)]{waskom_seaborn_2021}
Michael~L. Waskom.
\newblock seaborn: statistical data visualization.
\newblock \emph{Journal of Open Source Software}, 6\penalty0 (60):\penalty0 3021, 2021.
\newblock \doi{10.21105/joss.03021}.
\newblock URL \url{https://doi.org/10.21105/joss.03021}.
\newblock Publisher: The Open Journal.

\bibitem[Dietterich(1998)]{dietterich_approximate_1998}
Thomas~G. Dietterich.
\newblock Approximate {Statistical} {Tests} for {Comparing} {Supervised} {Classification} {Learning} {Algorithms}.
\newblock \emph{Neural Computation}, 10\penalty0 (7):\penalty0 1895--1923, October 1998.
\newblock ISSN 0899-7667, 1530-888X.
\newblock \doi{10.1162/089976698300017197}.
\newblock URL \url{https://direct.mit.edu/neco/article/10/7/1895-1923/6224}.

\end{thebibliography}

\section{Appendix}\label{sec6}
\begin{table}[ht]
\caption{Number of examples per finding for the ImaGenome dataset}
\centering
\begin{tabular}{lcc}
\textbf{Label} & \textbf{Yes} & \textbf{No} \\
\hline
Atelectasis & 140 & 310 \\
Pleural Effusion & 98 & 352\\
Pneumonia & 69 & 381 \\
Pneumothorax & 9 & 441 \\
\hline
\end{tabular}
\label{tab:findings_imagenome}
\end{table}

\begin{table}[ht]
\caption{Number of examples per finding for the Institutional dataset}
\centering
\begin{tabular}{lccc}
\textbf{Label} & \textbf{Yes} & \textbf{Maybe} & \textbf{No} \\
\hline
Atelectasis & 99 & 37 & 364 \\ 
Cardiomegaly & 55 & 1 & 444 \\ 
Consolidation & 36 & 4 & 460 \\ 
Edema & 46 & 6 & 448 \\ 
Lung Lesion & 27 & 15 & 458 \\ 
Lung Opacity & 245 & - & 255 \\ 
Pleural Other & 42 & 11 & 447 \\ 
Pleural Effusion & 78 & 18 & 404 \\ 
Pneumonia & 25 & 59 & 416 \\ 
Pneumothorax & 18 & 1 & 481 \\ 
Support Devices & 133 & - & 367 \\ 
Enl. Cardiomediastinum & 58 & 2 & 440 \\ 
Fracture & 50.0 & 3.0 & 447.0 \\ 
\hline
\end{tabular}
\label{tab:findings_institutional}
\end{table}

\end{document}